\documentclass[letterpaper, 10 pt, conference]{ieeeconf}
\usepackage{amsmath,amssymb,amsfonts}
\usepackage{orcidlink}
\usepackage{algorithmic}
\usepackage{float}
\usepackage{booktabs}
\usepackage{graphicx}
\graphicspath{{graphics/}}
\usepackage[list=true]{subcaption}
\usepackage{textcomp}
\usepackage{xcolor}
\def\BibTeX{{\rm B\kern-.05em{\sc i\kern-.025em b}\kern-.08em
    T\kern-.1667em\lower.7ex\hbox{E}\kern-.125emX}}
\usepackage[sorting=none]{biblatex}
\usepackage{hyperref}
\hypersetup{
  pdfauthor   = {},
  pdftitle    = {},
  pdfsubject  = {},
  pdfkeywords = {},
  pdfcreator  = {},
  pdfproducer = {}
}

\begin{document}

\title{
    A Lightweight Method for Validating Simulated LiDAR Configurations Based on Point Density over Distance
    }

\author{
Benedikt Degenstein\orcidlink{0009-0009-5413-2280},
Michael Spengler\orcidlink{0009-0008-7526-3565},
Julian Stähler\orcidlink{0000-0002-2672-5187},
and Carsten Markgraf\orcidlink{0000-0001-9447-2065}\\
\textit{Technical University of Applied Sciences Augsburg}\\
Augsburg, Germany\\
\{benedikt.degenstein, michael.spengler, julian.staehler, carsten.markgraf\}@tha.de
}

\maketitle

\begin{tikzpicture}[remember picture,overlay]
    \node[anchor=south,yshift=12pt] at (current page.south) {%
      \parbox{0.9\paperwidth}{\scriptsize\centering
        Accepted for publication at the 2026 IEEE 29th International Conference on
        Intelligent Transportation Systems (ITSC).
        \copyright~2026 IEEE. Personal use of this material is permitted. Permission
        from IEEE must be obtained for all other uses, in any current or future media,
        including reprinting/republishing this material for advertising or promotional
        purposes, creating new collective works, for resale or redistribution to servers
        or lists, or reuse of any copyrighted component of this work in other works.}};
  \end{tikzpicture}

\begin{abstract}
   Simulated LiDAR sensors play a crucial role in validating autonomous systems, yet achieving signal-level realism remains challenging due to simplified sensor models and limited configurability in common simulation environments. This paper presents a lightweight, platform-agnostic method to assess one observable aspect of LiDAR simulation fidelity: the range-dependent point density on annotated objects. The method uses only 3D bounding boxes and point clouds and does not require semantic labels, detailed maps, simulator internals, or perception models. To quantify agreement between real and simulated range-dependent point distributions, we formulate the Density Similarity Score (DSS), a bounded similarity representation derived from the Mean Absolute Percentage Error (MAPE) over discretized distance bins. The method is evaluated against a real-world LiDAR using three CARLA configurations of varying complexity. Results show that naive parameter matching based on sensor datasheets leads to systematic point-density deviations, particularly in the near field, while a custom configuration using the sensor's non-uniform vertical beam angles achieves the highest similarity in the evaluated scenario (DSS: 0.559--0.709). The proposed approach is intended as an early-stage, signal-level consistency check for identifying gross mismatches in simulated LiDAR configurations. It should be interpreted as a complementary validation step rather than a complete measure of simulation quality or downstream perception performance.
\end{abstract}

\section{Introduction}
\begin{figure}[]
    \centering
    \includegraphics[width=0.35\textwidth]{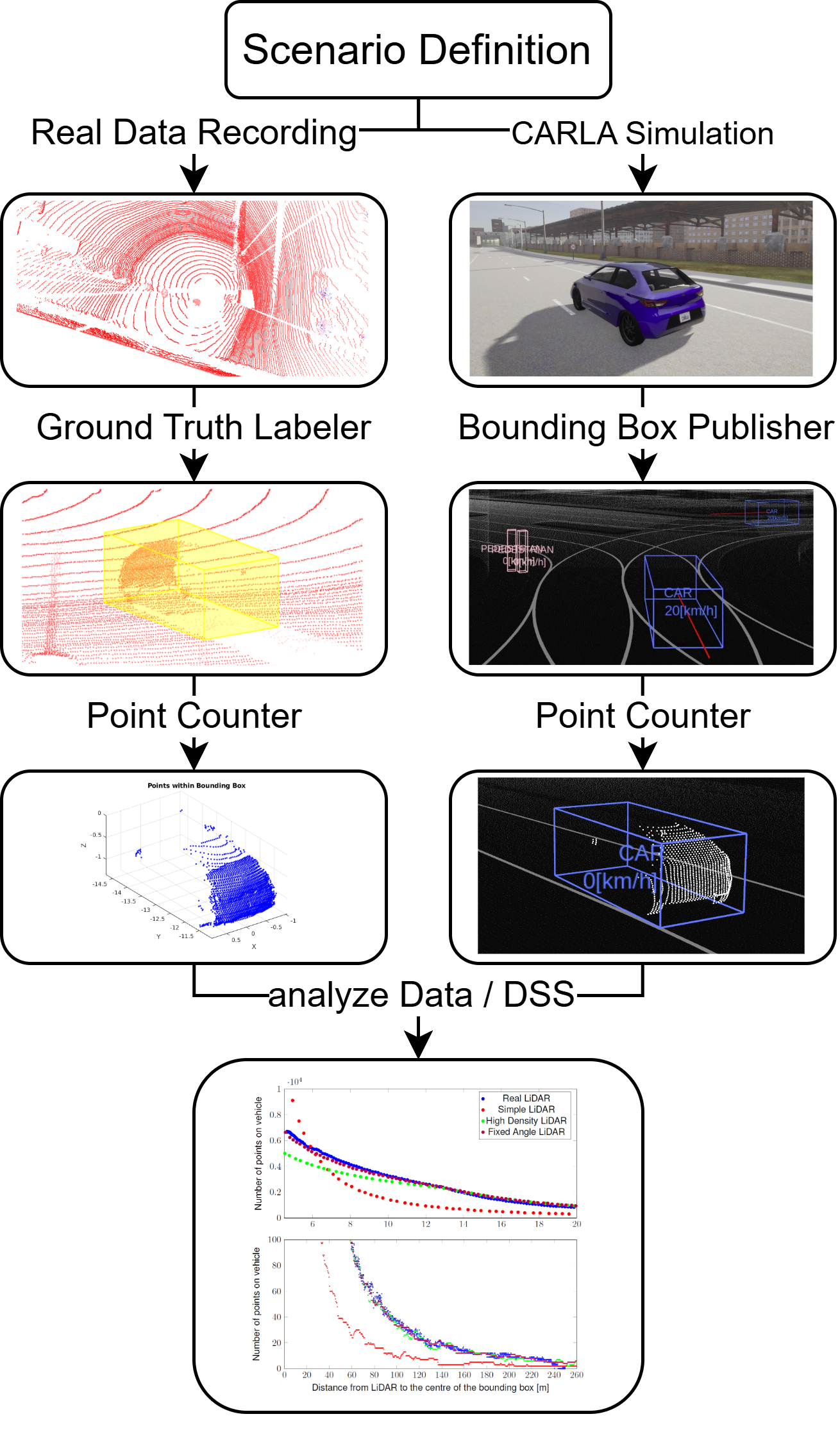}
    \caption{Overview of the proposed validation pipeline. Both the real-world and simulation-based paths originate from the same scenario definition, ensuring structural comparability of the resulting density-over-distance profiles.}
    \label{fig:flowchart}
\end{figure}
Simulation plays an increasingly vital role in the development and validation of autonomous systems \parencite{yurtseverSurveyAutonomousDriving2020}. It enables scalable, repeatable, and cost-efficient testing under controlled conditions \parencite{riedmaierSurveyScenarioBasedSafety2020}, particularly important when working with safety-critical functions such as object detection, tracking, or collision avoidance. Among the most commonly used sensors in autonomous vehicles, LiDAR stands out for its ability to provide high-resolution 3D representations of the environment \parencite{royoOverviewLidarImaging2019}.

To enable realistic evaluation of perception systems, simulation environments must offer sensor models that closely approximate real-world behavior. However, achieving such realism for LiDAR remains challenging: many simulators provide limited configurability or abstract approximations of sensor geometry, and existing validation methods often operate at the task level, using perception performance or high-level metrics as indirect measures of signal fidelity. This leaves a gap for low-level validation techniques that directly assess the structure of simulated LiDAR signals, such as point density over distance, instead of only evaluating downstream tasks.

This paper introduces a simple, platform-agnostic method to evaluate how closely a simulated LiDAR reproduces the range-dependent object return density observed with a real sensor. The approach focuses on analyzing point density within object bounding boxes as a function of distance, yielding density-over-distance profiles that can be compared between real and simulated data without requiring semantic segmentation, detailed maps, simulator internals, or trained perception models. In contrast to prior work that validates simulators via 3D object-level comparison or autonomy-stack performance \parencite{wallaceValidatingSimulationEnvironments2022,schaermannValidationVehicleEnvironment2017,manivasagamLiDARsimRealisticLiDAR2020,manivasagamZeroDomainGap2023}, the proposed method targets a specific signal-level property of the raw point cloud: object point count as a function of range. To demonstrate the method, we apply it to a real world LiDAR \parencite{RubyRoboSenseSafer} and three different simulated configurations in CARLA \parencite{dosovitskiyCARLAOpenUrban2017}.
The main contributions of this paper are:
\begin{itemize}
    \item A platform-agnostic, model-free validation method based on 
          density-over-distance analysis of LiDAR point clouds within 
          object bounding boxes.
    \item The Density Similarity Score (DSS), a bounded MAPE-derived score for summarizing
          agreement between real and simulated range-dependent point-count profiles
    \item An empirical evaluation across three CARLA configurations 
          against a real world LiDAR, showing that non-uniform vertical beam modeling substantially improves range-dependent point-density agreement in the evaluated CARLA setup.
  \end{itemize}
We address the question of how to efficiently validate whether a simulated LiDAR configuration reproduces the distance-dependent point density of a real sensor. To quantify agreement between real and simulated density profiles, we introduce the Density Similarity Score (DSS). Experiments with CARLA configurations and a real world LiDAR show that the configuration using the sensor's non-uniform vertical beam angles achieves the highest DSS overall, while the naive datasheet-based configuration performs worst.

\section{Related Works}
The validation of simulated LiDAR data remains an important research topic for ensuring the reliability of autonomous systems in safety-critical scenarios \parencite{wallaceValidatingSimulationEnvironments2022,manivasagamLiDARsimRealisticLiDAR2020}. Many works have addressed the realism of LiDAR simulations, the mitigation of the sim-to-real domain gap, and the evaluation of sensor fidelity. However, prior studies \parencite{schaermannValidationVehicleEnvironment2017,manivasagamZeroDomainGap2023,aliComprehensiveAssessmentLiDAR2025} focused on perception-level validation or employed high-level task performance as a proxy for simulation accuracy, leaving a gap for lightweight, interpretable low-level signal validation strategies.

Wallace et al. \parencite{wallaceValidatingSimulationEnvironments2022} proposed a simulation validation methodology using a 3D object dissimilarity metric between real and simulated LiDAR point clouds in matched driving scenarios. Their work underlines the importance of directly comparing geometric sensor outputs to assess simulation fidelity but primarily evaluates object-level shape similarity rather than distance-dependent point density. Similarly, Schaermann et al. \parencite{schaermannValidationVehicleEnvironment2017} introduced a two-stage validation framework where real and simulated sensor data are first compared at the raw point cloud level, followed by downstream Advanced Driver Assistance Systems module performance comparison. This structured approach supports the notion that early-stage discrepancies in LiDAR signal structure can propagate into higher-level failures, yet it does not provide a simple, stand-alone metric for characterizing density-over-distance behavior.

Ali et al. \parencite{aliComprehensiveAssessmentLiDAR2025} conducted a comprehensive comparison of LiDAR evaluation metrics using both simulated and real data. They showed that commonly used metrics such as the Chamfer Distance (CD) exhibit only moderate sensitivity to variations in point cloud density, especially as overall density increases, and that density-aware adaptations can improve robustness. These findings emphasize the importance of accounting for density distribution in LiDAR evaluation but focus on general-purpose distance metrics rather than a dedicated, interpretable density-over-distance profile.

Another line of research aims to improve or analyze the realism of LiDAR simulation itself. Manivasagam et al. \parencite{manivasagamLiDARsimRealisticLiDAR2020} presented LiDARsim, a hybrid physics- and learning-based simulator for autonomous-driving applications such as vehicle segmentation and detection, simulated-data augmentation, rare-object perception testing, and safety-critical autonomy testing. Their later paired-scenario study \parencite{manivasagamZeroDomainGap2023} analyzes how LiDAR simulation domain gap affects detection, motion forecasting, and motion planning. Importantly, their findings indicate that LiDAR simulation realism is multi-factorial, involving scanning and geometric effects such as calibrated intrinsics, actor geometry, rolling shutter, and motion blur, as well as return-level effects such as raydrop, material response, spurious points, and multi-echo returns. Complementary to this, Zhao et al. \parencite{zhaoMethodApplicationsLidar2021} proposed a LiDAR modeling method incorporating physical phenomena such as atmospheric attenuation and surface reflectance.

In contrast, our work does not aim to close the full LiDAR sim-to-real gap or to predict autonomy-level performance. Instead, it provides a lightweight diagnostic for one specific signal-level observable: range-dependent object return density. Using only point clouds and 3D bounding boxes, DSS summarizes how closely a simulated configuration reproduces the real sensor's distance-dependent object point-count profile. The method is intended as an early-stage consistency check that complements, rather than replaces, point-set metrics and downstream autonomy
validation.
\section{Method}\label{sec:method}
\subsection{Motivation and Sensor Modeling Challenges}
Simulated LiDAR configurations in CARLA are often parameterized from manufacturer datasheets (e.g., channel count, vertical field of view (FOV), and points per second). However, matching these high-level parameters is insufficient to reproduce the range-dependent point distribution of real sensors. CARLA assigns vertical beams uniformly within the specified FOV, whereas real LiDAR sensors employ non-uniform vertical beam layouts with higher beam density near the horizon and sparser spacing toward the FOV limits. As a result, datasheet-level parameter matching can yield systematic discrepancies in point density over distance, even when nominal specifications appear consistent. This motivates the need for a lightweight, platform-agnostic signal-level validation procedure that can assess range-dependent differences between simulated and real LiDAR returns.

\subsection{Density-over-Distance Validation Approach}
We analyze LiDAR signal structure by plotting the number of returns within an object's 3D bounding box as a function of the object's range to the LiDAR sensor. For spinning multi-beam LiDARs with repeatable angular sampling, the angular sampling pattern induces a characteristic range-dependent point distribution. Mismatches in beam layout, field of view, rendering assumptions, or return modeling can therefore manifest as systematic deviations in these point-count-versus-range patterns.

Operationally, the method tracks a target object over a range of distances and counts the LiDAR points within its 3D bounding box at each time step. The resulting (range, point count) samples are visualized to qualitatively compare real and simulated configurations.

The approach provides an interpretable signal-level comparison that uses only point clouds and 3D bounding boxes, without relying on perception models or task-level metrics.
Since such profile comparisons are qualitative, we introduce a compact metric to summarize the similarity of point distributions in a single score. Fig. \ref{fig:flowchart} illustrates the full validation pipeline, covering both the real-world and simulation-based processing paths.

\subsection{Density Similarity Score}
To enable a quantitative comparison of simulated and real LiDAR point-count patterns across sensor configurations, we adapt the well-established Mean Absolute Percentage Error (MAPE) into a bounded similarity score tailored to density-over-distance profiles. Rather than reporting raw error magnitudes, we invert and clip the MAPE to yield a score in $[0,1]$, which we refer to as the Density Similarity Score (DSS). MAPE is commonly reported in percent by multiplying by 100, but we omit this scaling and work with the unitless form.

We discretize object range $d$ (in meters) into bins of width $\Delta d$. In this study, we use $\Delta d = 1\,\mathrm{m}$ as a practical default. The bin index is given by $i(d) = \lfloor d/\Delta d \rfloor$, and bin $i$ corresponds to the interval $[i\Delta d,(i+1)\Delta d)$. For each bin $i$, let $T_i$ denote the number of frames whose object range falls into that bin. The per-bin mean point count for the simulated and real data is then given by
\begin{equation}
    s_i = \frac{1}{T_i} \sum_{t=1}^{T_i} s_{i,t},
\end{equation}
\begin{equation}
    r_i = \frac{1}{T_i} \sum_{t=1}^{T_i} r_{i,t},
\end{equation}
where \(s_{i,t}\) and \(r_{i,t}\) are the simulated and real point counts within bin \(i\) at time \(t\), respectively. Bins with \(r_i = 0\) are excluded to avoid division by zero in the relative error and because they do not contribute information about density agreement at that distance. Let \(\mathcal{I} = \{ i \mid r_i > 0 \}\) denote the set of remaining bins and \(N = |\mathcal{I}|\).

For each \(i \in \mathcal{I}\), we compute the absolute relative error
\begin{equation}
    e_i = \left| \frac{s_i - r_i}{r_i} \right|.
\end{equation}
Averaging \(e_i\) over \(i \in \mathcal{I}\) yields the unitless MAPE over distance bins. To obtain a similarity-style score for ranking and diagnosing configurations, we define
\begin{equation}
    \mathrm{DSS} = \max\left(0,\, 1 - \frac{1}{N} \sum_{i \in \mathcal{I}} e_i \right),
\end{equation}
which ensures \(\mathrm{DSS} \in [0,1]\). A DSS value of 1 indicates perfect agreement under this criterion, while DSS = 0 indicates an average absolute relative deviation of at least 1.0, i.e., 100\%, across the considered distance bins and therefore flags a severe mismatch in the evaluated density profile.

Since DSS is computed solely from geometric point counts within 3D bounding boxes, it is independent of perception models and does not require access to simulator internals. DSS is used in the experimental evaluation to compare LiDAR configurations with respect to range-dependent object point-density agreement with the real sensor reference.

\section{Experimental Setup and Results}\label{sec:setup_results}
To evaluate the proposed validation method, we collected scenario-analogous LiDAR datasets in simulation and in a controlled real-world setup. Recordings from a real LiDAR sensor serve as the reference, against which different CARLA LiDAR configurations are assessed by comparing their range-dependent point-count behavior.

To reduce confounding effects from scenario complexity and make sensor-configuration effects more visible, we selected a simple and reproducible dynamic scenario that can be implemented in both domains with minimal effort. This allowed us to focus on sensor-configuration effects while limiting, but not eliminating, scenario-related confounders. In both the pedestrian and vehicle scenarios, the dynamic actor moved away from the ego vehicle along a straight path at approximately walking speed (about 1-1.5\,m/s) in simulation and in the real world. For each combination of actor type (pedestrian, vehicle) and domain (real, simulated configuration), one representative run was selected for analysis.
An overview of the experimental procedure is shown in Fig. \ref{fig:flowchart}. While the simulation and real-world data collection were performed independently and at separate times, both follow structurally analogous pipelines: scenario execution, bounding box acquisition, and point counting, ultimately producing comparable datasets for density-over-distance analysis. The following sections describe each part of this process in detail and present the resulting comparisons.

\subsection{Simulation Setup}\label{sec:sim_setup}
The simulation experiments were conducted using the CARLA simulator (version 0.9.13) with Robot Operating System (ROS) integration via the carla\_ros\_bridge \parencite{CarlasimulatorRosbridge2025}. A stationary ego vehicle was equipped with a virtual LiDAR sensor mounted at roof height, mimicking the setup used in the real-world experiments. While the real sensor is a Robosense Ruby Plus, CARLA's default LiDAR implementation uses a raycasting-based sensor model that requires manual configuration to approximate real sensor behavior \parencite{SensorsReferenceCARLA}.

The real-world reference sensor and the three simulated configurations are summarized below. The RoboSense Ruby Plus \parencite{RubyRoboSenseSafer} serves as the empirical reference against which all configurations are evaluated:

\begin{itemize}
  \item \textbf{RoboSense Ruby Plus (Real Sensor):} 128 channels, ${\sim}$4,608,000 points per second (Balance mode, dual return), vertical FoV from +15° to -25°, maximum range 250\,m. Crucially, beam angles are \textit{non-uniformly} distributed: 104 channels are concentrated in the central zone (+3.81° to -6.51°) at up to 0.1° vertical resolution, with sparser spacing toward the FoV limits.

  \item \textbf{Simple:} 128 channels, 4,608,000 points per second, vertical field of view from +15° to -25°, with \textit{uniformly} distributed vertical beam angles.

  \item \textbf{High Density:} 104 channels, 3,744,000 points per second, vertical field of view from +3.81° to -6.51°, also with uniform vertical spacing.

  \item \textbf{Custom:} 128 channels, 4,608,000 points per second, with vertical beam angles individually set according to the RoboSense Ruby Plus specification by modifying the CARLA Unreal Engine plugin source code to support custom vertical angle assignments for each laser beam.
\end{itemize}
For each configuration, we executed the scenario described in Section \ref{sec:setup_results}, using a straight-line trajectory up to 250\,m (the LiDAR's configured maximum range). A custom ROS node published the actor's ground-truth bounding box in real time. Simultaneously, a point counting node computed the number of LiDAR returns falling within the bounding box at each frame. The object's distance to the sensor and the corresponding number of detected points were logged for further analysis.

\subsection{Real-World Data Collection}
\begin{figure}
    \centering
    \includegraphics[width=0.48\textwidth]{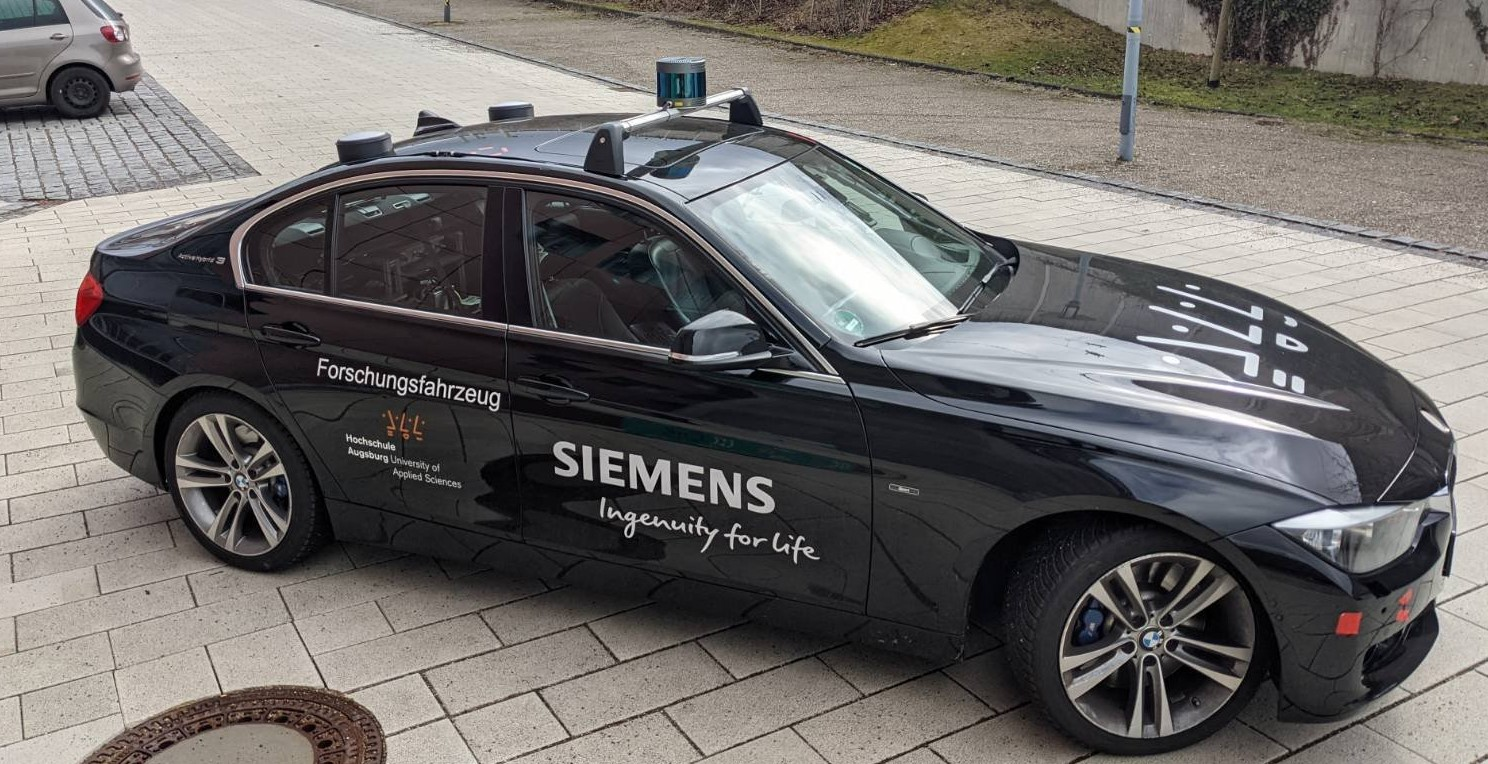}
    \caption{Research vehicle used for real-world data collection, equipped with a rooftop sensor mount and a RoboSense Ruby Plus LiDAR sensor.}
    \label{fig:bmw}
\end{figure}

To enable a meaningful comparison with the simulation, the same scenario was reproduced in a controlled real-world environment. The experimental setup consisted of a stationary ego vehicle equipped with a RoboSense Ruby Plus LiDAR sensor mounted on the roof via a dedicated rooftop carrier, as shown in Fig. \ref{fig:bmw}.
The real-world runs followed the same straight-line scenario described in Section \ref{sec:setup_results}, with the actor moving away from the stationary ego vehicle.

The data was recorded on a flat, open environment with minimal occlusions or background clutter. This location allowed for consistent and undisturbed LiDAR returns over long distances, closely matching the clean conditions used in simulation.

All LiDAR frames and associated timestamps were captured using ROS. Real-time bounding box annotations were not available during data collection. Therefore, the annotation was performed offline using MATLAB's Ground Truth Labeler tool \parencite{GroundTruthLabeler}. Once the bounding boxes are generated, the same point counting algorithm used in the simulation is applied to the real-world data. For every annotated frame, the number of LiDAR points falling within the bounding box was computed, and the corresponding distance between the object and the sensor was extracted. The resulting per-frame point counts and object distances were stored for subsequent analysis.

\subsection{Data Processing and Binning Strategy}
For both real and simulated data, the object distance was discretized using the same binning strategy across all configurations. We use a default bin width of $\Delta d = 1\,\mathrm{m}$ from 0 m up to 250 m. This value was chosen as a practical trade-off between profile resolution and per-bin sample aggregation. It should be interpreted as an experimental setting rather than a universal constant; evaluating DSS sensitivity to alternative bin widths is left for future work.

\subsection{Results and Comparison}
\begin{figure*}[ht]
    \centering
    \begin{subfigure}[b]{0.24\textwidth}
        \includegraphics[width=\textwidth]{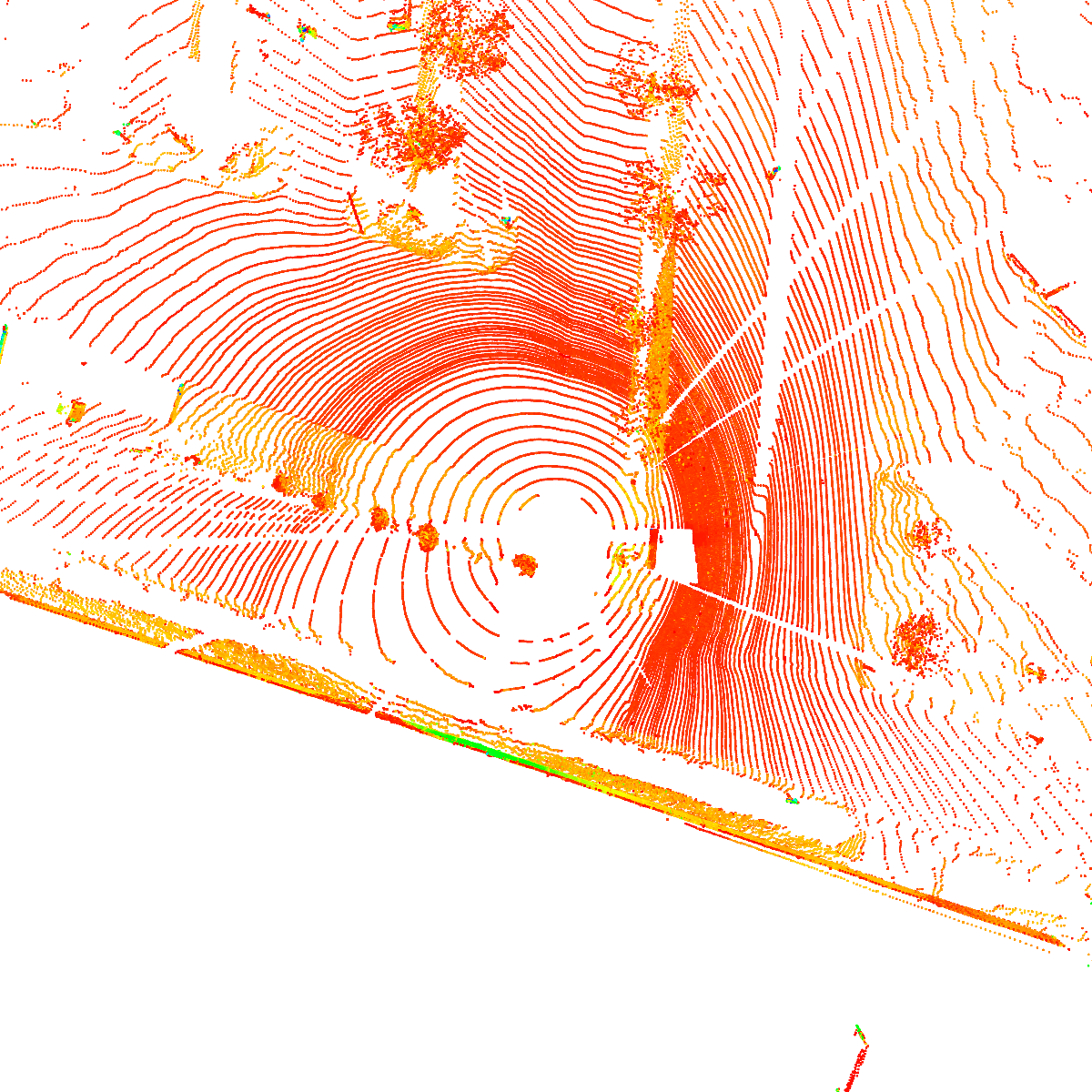}
        \caption{Real-world LiDAR}
        \label{subfig:real_lidar}
    \end{subfigure}
    \hfill
    \begin{subfigure}[b]{0.24\textwidth}
        \includegraphics[width=\textwidth]{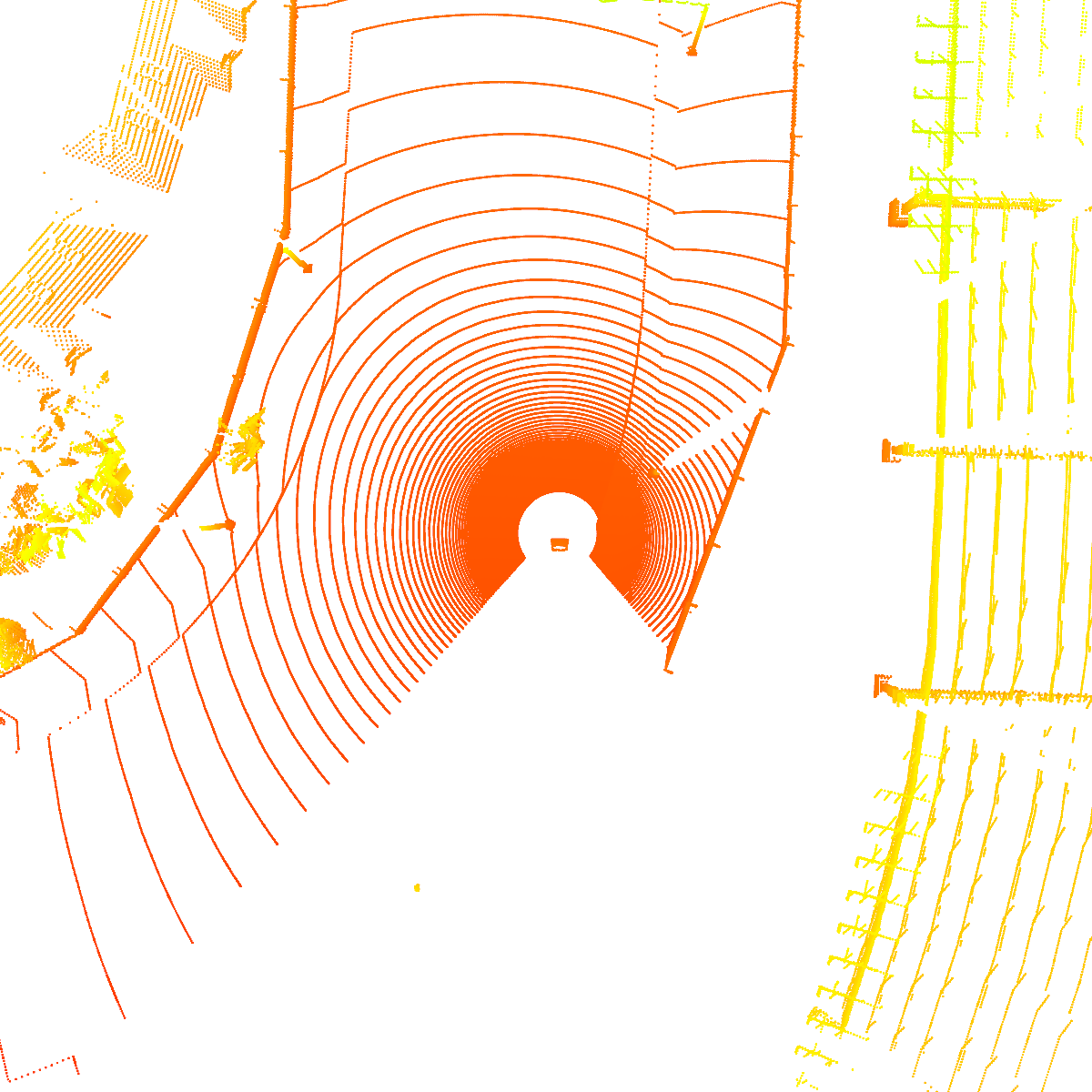}
        \caption{Simple config}
        \label{subfig:simple_lidar}
    \end{subfigure}
    \hfill
    \begin{subfigure}[b]{0.24\textwidth}
        \includegraphics[width=\textwidth]{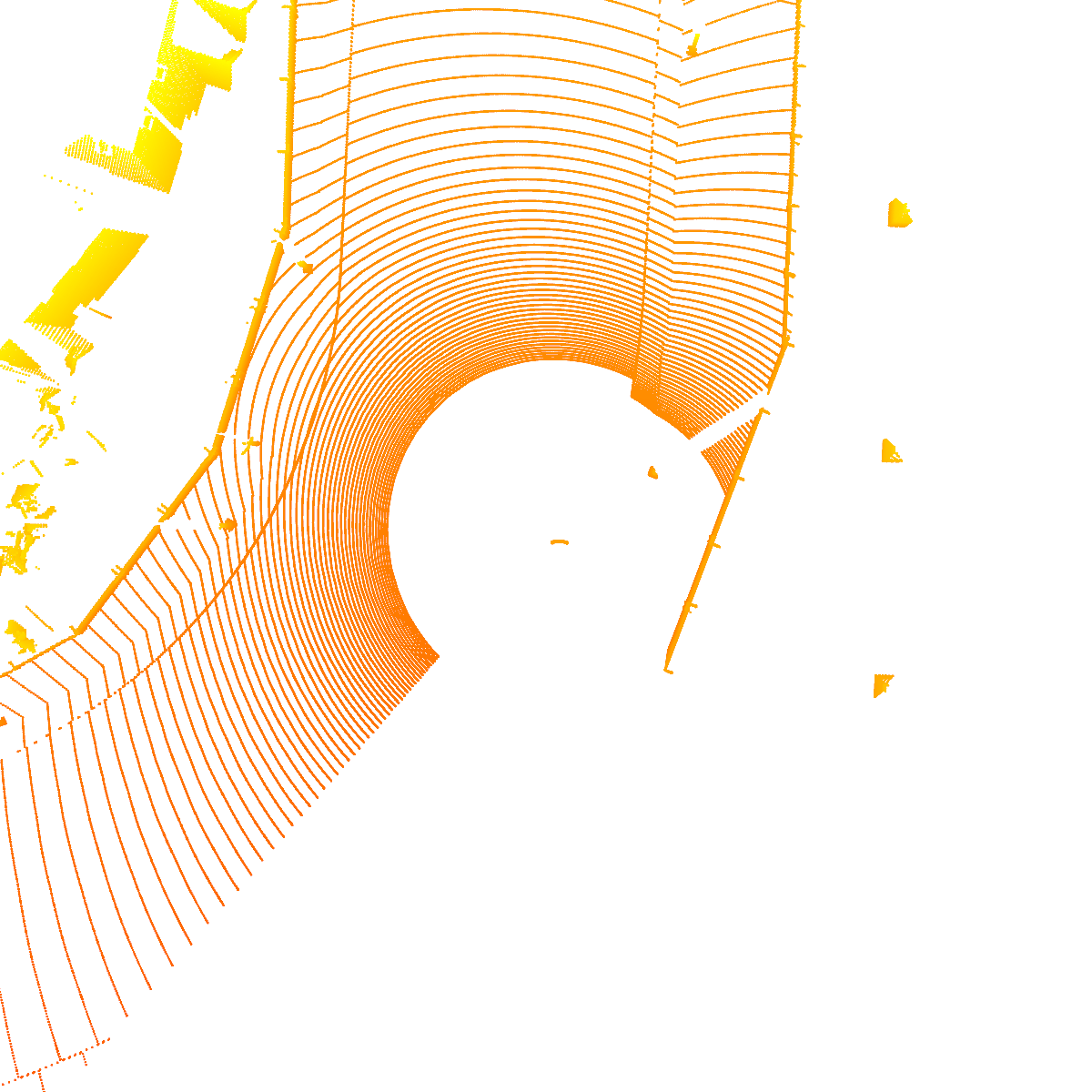}
        \caption{High-density config}
        \label{subfig:high-dens_lidar}
    \end{subfigure}
    \hfill
    \begin{subfigure}[b]{0.24\textwidth}
        \includegraphics[width=\textwidth]{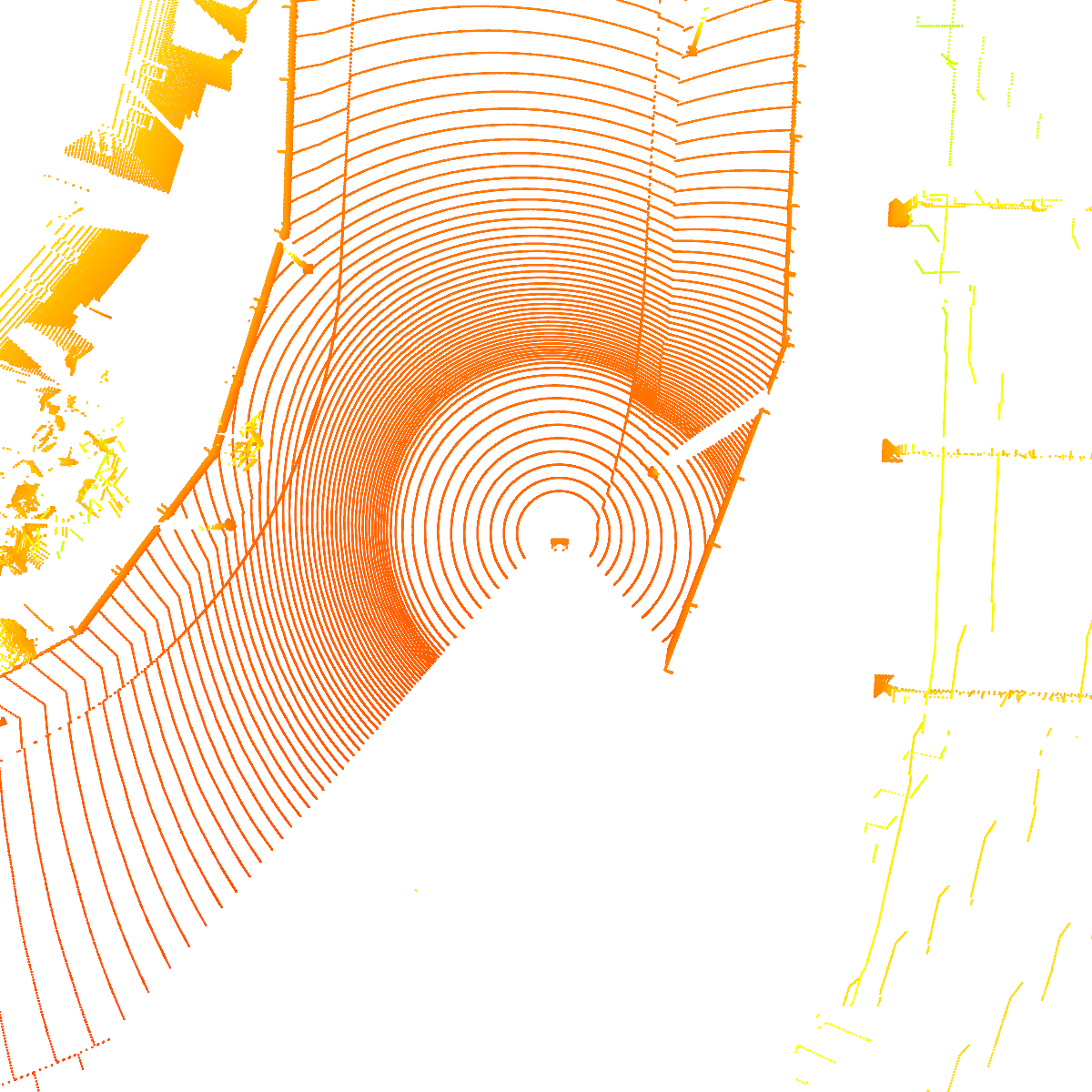}
        \caption{Custom angle config}
        \label{subfig:custom_angle_lidar}
    \end{subfigure}
    \caption{Illustrative top-down views of representative LiDAR point clouds for the real sensor and the three simulated configurations. The full-scene views are not used for quantitative comparison. DSS is computed from object-centric point counts within 3D bounding boxes.}
    \label{fig:lidar_comparison}
\end{figure*}

Fig. \ref{fig:lidar_comparison} provides illustrative full-scene point-cloud examples for the real sensor and the three simulated configurations. These views are included only to illustrate qualitative differences in point-cloud structure, such as ring patterns and beam spread. They are not used for quantitative evaluation, and scene-level visual similarity is not claimed because the real and simulated backgrounds, occlusions, and object layouts are not identical. The quantitative comparison is based solely on object-centric point counts within 3D bounding boxes, as shown in Fig. 4 and summarized by DSS.

\begin{figure*}[t]
    \centering
    \begin{subfigure}[b]{0.495\textwidth}
        \includegraphics[width=\linewidth]{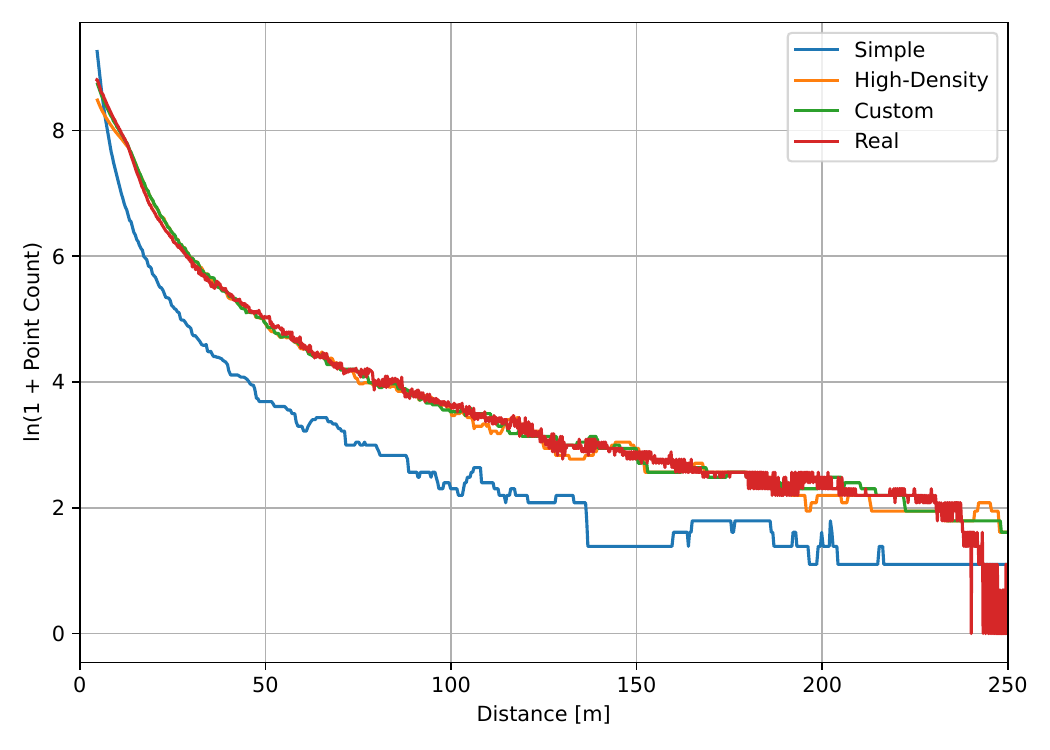}
        \caption{Vehicle scenario: full range (log-scaled)}
        \label{fig:vehicle_log}
    \end{subfigure}
    \hfill
    \begin{subfigure}[b]{0.495\textwidth}
        \includegraphics[width=\linewidth]{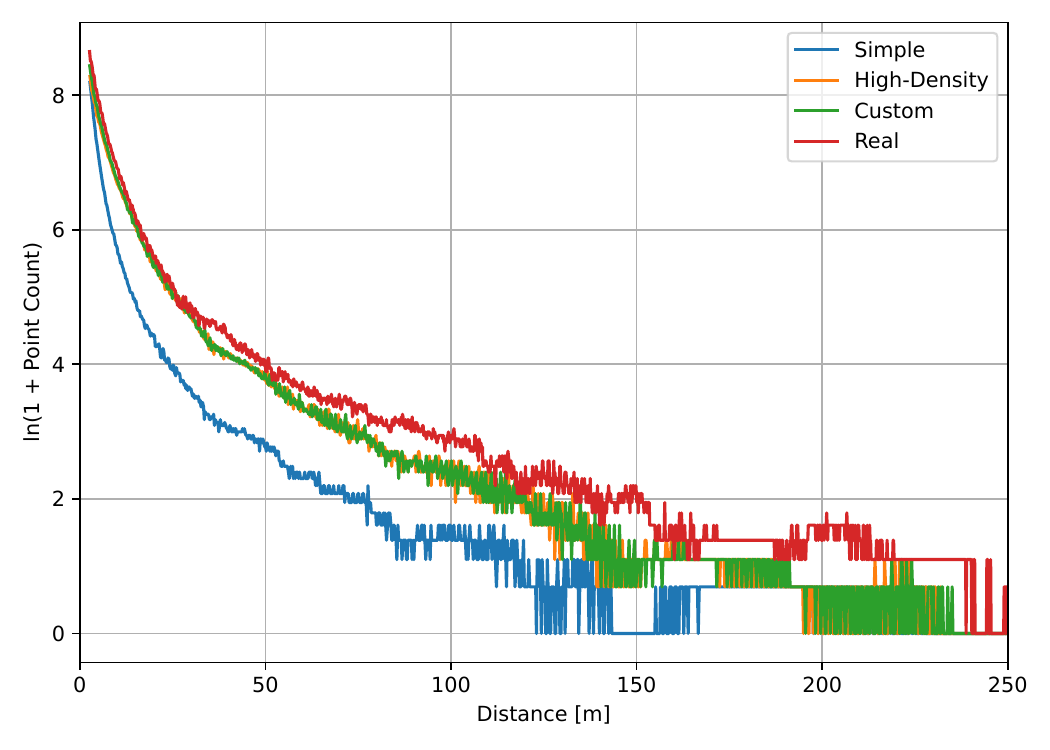}
        \caption{Pedestrian scenario: full range (log-scaled)}
        \label{fig:pedestrian_log}
    \end{subfigure}

    \begin{subfigure}[b]{0.495\textwidth}
        \includegraphics[width=\linewidth]{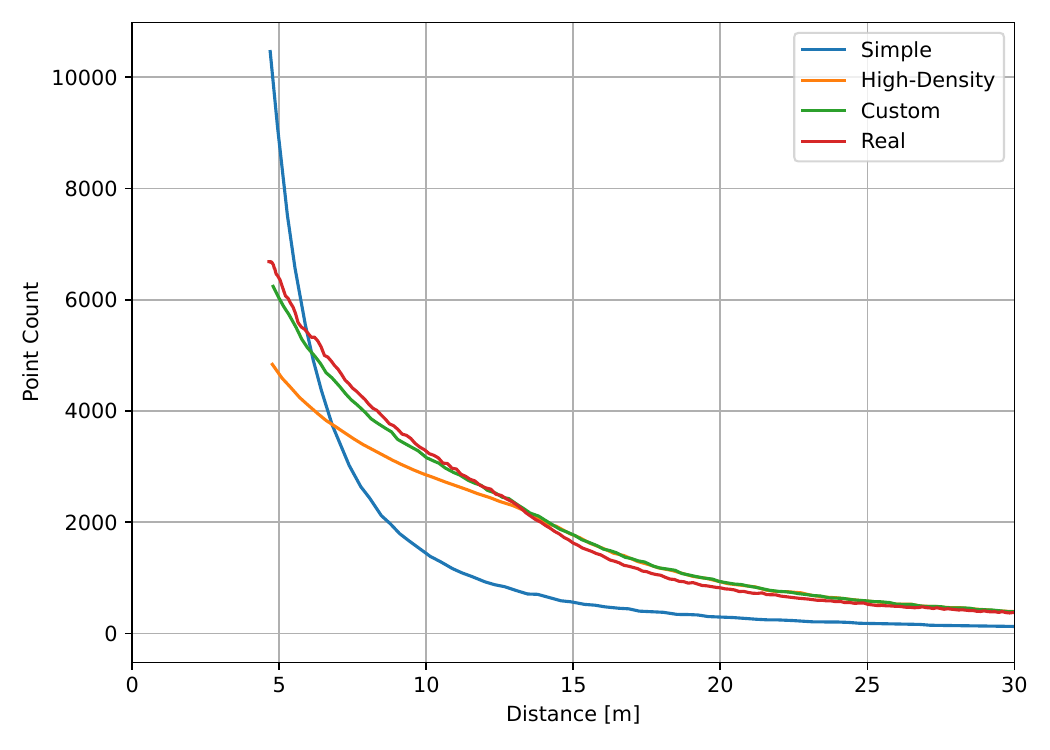}
        \caption{Vehicle scenario: near-field (linear, 0-30\,m)}
        \label{fig:vehicle_zoom}
    \end{subfigure}
    \hfill
    \begin{subfigure}[b]{0.495\textwidth}
        \includegraphics[width=\linewidth]{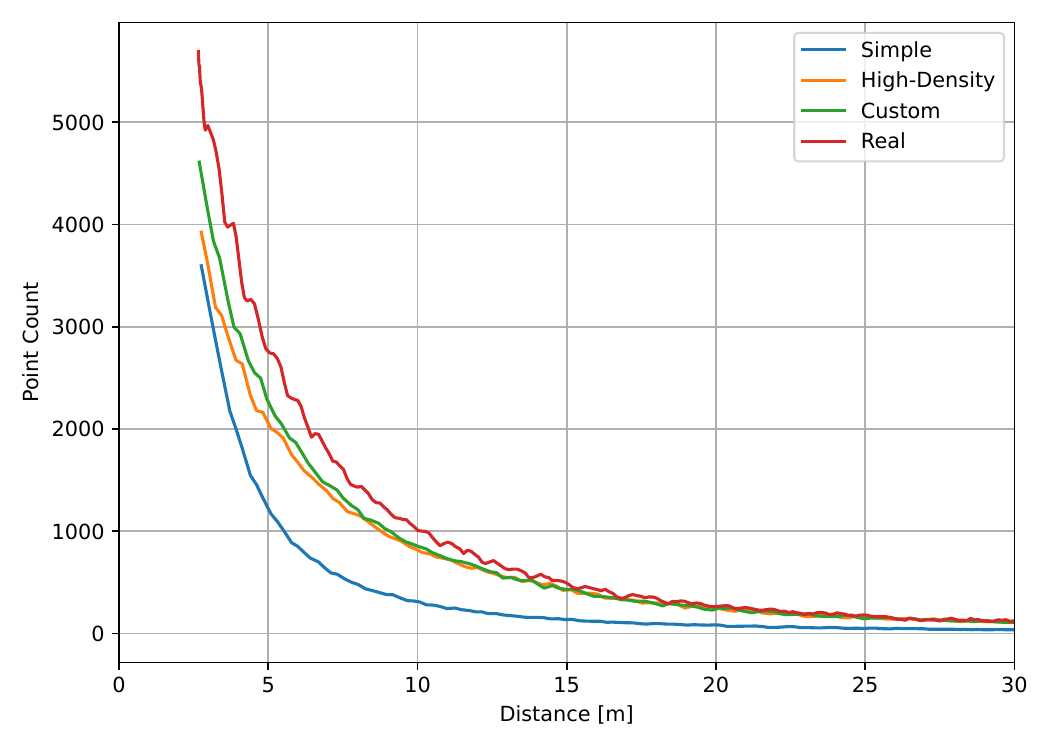}
        \caption{Pedestrian scenario: near-field (linear, 0-30\,m)}
        \label{fig:pedestrian_zoom}
    \end{subfigure}
    \caption{Comparison of LiDAR point counts across distance for all configurations. 
    The top row shows the full distance range (logarithmically scaled to emphasize long-range behavior), 
    while the bottom row provides a linear view of the near-field region up to 30 meters.}
    \label{fig:lidar_point_distribution}
\end{figure*}

Fig. \ref{fig:lidar_point_distribution} shows the results of the density-over-distance validation approach described in Section \ref{sec:sim_setup}. Each configuration's number of LiDAR points within the bounding box is plotted against the object's distance from the sensor. The top row provides a logarithmically scaled overview across the full sensing range, while the bottom row zooms into the near-field region up to 30 meters using a linear scale to emphasize close-range behavior.

In the vehicle scenario (Figs. \ref{fig:vehicle_log} and \ref{fig:vehicle_zoom}), the Simple configuration shows the largest deviation from the real density profile under the proposed criterion: significantly higher point counts in the near field ($<$30 m) and a rapid drop-off, lacking the characteristic inflection at ~13 m observed in the real sensor. This systematic deviation is due to the uniform vertical beam distribution, which fails to replicate the real sensor's non-uniform geometry. The High-Density configuration approaches the real-sensor profile only beyond 13 m but exhibits poor near-field coverage due to its limited vertical FOV. The Custom configuration aligns most closely with the real device across the full range.

In the pedestrian scenario (Figs. \ref{fig:pedestrian_log} and \ref{fig:pedestrian_zoom}), all simulated configurations show similar overall trends but consistently lower point counts than the real LiDAR. The Simple configuration produces a higher point density near the sensor but drops sharply with distance. High-Density underperforms at short ranges due to its limited vertical FOV, while Custom achieves the best overall approximation. Some deviations likely stem from model differences between the simulated and real pedestrian, complicating the definition of strict acceptability thresholds.

These findings indicate that vertical beam layout can strongly influence object point-density profiles and should be considered when the intended use case depends on realistic object return
density.

\subsection{Quantitative Evaluation via Density Similarity Score}
While the point distribution plots offer valuable qualitative insights, they can be difficult to interpret when comparing multiple configurations simultaneously. The Density Similarity Score (DSS) provides a simple and interpretable metric to quantify the similarity between simulated and real LiDAR point distributions.

Table~\ref{tab:dss_results} summarizes the DSS values for each configuration and object type. Higher values indicate greater similarity to the real sensor's distance-binned object point-count profile.

\begin{table}[H]
\centering
\caption{Density Similarity Score (DSS) for Each Configuration}
\label{tab:dss_results}
\begin{tabular}{lcc}
\toprule
\textbf{Configuration} & \textbf{Pedestrian DSS} & \textbf{Vehicle DSS} \\
\midrule
Custom LiDAR           & 0.559             & 0.709               \\
High-Density LiDAR     & 0.558             & 0.656               \\
Simple LiDAR           & 0.213             & 0.267               \\
\bottomrule
\end{tabular}
\end{table}

The DSS results are consistent with the density-over-distance profiles in Fig. 4. The Custom LiDAR achieves the highest overall similarity to the real sensor's distance-binned object point-count profile (DSS = 0.559/0.709 for pedestrian/vehicle), followed by High-Density (0.558/0.656). The nearly identical pedestrian DSS values for Custom and High-Density indicate that configuration ranking can depend on object type and scenario details. In contrast, the Simple configuration shows the largest deviation (0.213/0.267), consistent with the mismatch introduced by its uniform beam layout. These values indicate relative agreement under the proposed density-over-distance criterion. Whether a given DSS value is sufficient depends on the intended downstream use case and requires additional task-level validation.

\section{Conclusion and Discussion}
The results indicate that, for the controlled scenarios considered in this study, vertical beam geometry is a major contributor to range-dependent object point-density agreement in the evaluated CARLA setup. The Custom configuration achieves the closest agreement with the real sensor's distance-binned object point-count profile, while the naive datasheet-based Simple configuration shows the largest deviations, particularly in the near field. These findings show that lightweight, model-free density analysis over distance can help identify clearly mismatched simulated LiDAR configurations under controlled conditions. However, DSS should be interpreted as a diagnostic indicator for one specific signal-level observable rather than as a complete measure of LiDAR simulation quality.

The current study focuses exclusively on geometric point density as a similarity signal, deliberately trading completeness for interpretability and portability. This scope is a natural starting point, but future extensions could incorporate intensity distributions, noise profiles, and multi-echo return rates into a vector-valued similarity measure, offering a richer characterization of sensor behavior without abandoning the lightweight, model-free design principle. Scenarios were intentionally kept simple, straight-line motion on flat terrain with minimal occlusions, which limits direct generalizability to complex urban environments or sloped surfaces. However, this also makes the method a well-controlled baseline against which more expressive scenario databases can be validated in follow-up work. The reliance on a single run per configuration leaves statistical variability unquantified, an issue that Monte Carlo sampling across scenario variants could systematically address while simultaneously stress-testing DSS stability. Motion distortion effects are expected to be small in this study, as the ego vehicle is stationary and the dynamic actor moves at low speed, resulting in displacement per LiDAR scan that is small relative to the chosen default bin width. Perhaps most concretely, the manual offline bounding box annotation of real-world data introduces human labeling bias as a potential confound. Automated object detection pipelines or established annotation tools could replace this step and the density profiles generated here could themselves serve as a low-cost benchmark for evaluating the geometric consistency of such auto-labeling algorithms in sparse, long-range point clouds. Quantifying the relationship between DSS and downstream perception performance, e.g., by correlating DSS values with detection recall across distance bins, remains an important direction for future work and would further ground DSS interpretation in task-level evidence.

The DSS-based approach presented here contributes a practical first-pass consistency check for the sensor modeling cycle. It can be applied before perception models are trained or large scenario libraries are scaled up, where gross configuration mismatches are comparatively cheap to correct. A low DSS value should be interpreted as evidence of a mismatch in range-dependent object return density; possible causes include beam geometry, field of view, object geometry, annotation quality, reflectivity assumptions, and return modeling. In the CARLA/RoboSense Ruby Plus case studied here, revisiting the vertical beam angle assignment was an effective corrective action. More generally, DSS is intended to complement, not replace, high-fidelity physics simulation, point-set metrics, and downstream validation with perception, prediction, or planning modules.

\section*{Acknowledgment}
This work was funded by the German Federal Ministry for Economic Affairs and Climate Action (BMWK) under grant no. 19A21051K, "Real Driving Validation (RDV)" (2022-2025). The authors thank their colleagues at TTZ Landsberg for their continuous support, valuable advice, and contributions throughout the project.

\printbibliography
\end{document}